\documentclass[10pt,twocolumn,letterpaper]{article}

\usepackage{cvpr}              

\usepackage{graphicx}
\usepackage{amsmath}
\usepackage{amssymb}
\usepackage{booktabs}
\usepackage{algorithm}
\usepackage{algorithmicx}
\usepackage{algpseudocode}
\usepackage{multirow}
\usepackage[table,xcdraw]{xcolor}
\usepackage{booktabs}
\usepackage{verbatim}
\usepackage{color, colortbl}
\usepackage{bm}
\usepackage{enumitem}
\usepackage{pifont}
\usepackage[accsupp]{axessibility}
\usepackage{multirow}
\usepackage{makecell}
\usepackage{adjustbox}
\usepackage{bbding}
\usepackage{tikz}
\usepackage{pgfplots}
\usepackage{ulem}
\usepackage{mathtools}
\usepackage{pgfplotstable}
\usepackage{setspace}
\usepackage{caption}
\usepackage{amsmath}

\definecolor{cvprblue}{rgb}{0.21,0.49,0.74}
\usepackage[pagebackref,breaklinks,colorlinks,allcolors=cvprblue]{hyperref}
\usepackage[capitalize]{cleveref}
\Crefname{section}{Section}{Sections}
\Crefname{table}{Table}{Tables}
\Crefname{figure}{Figure}{Figures}
\def\paperID{} 
\def\confName{CVPR}
\def\confYear{2026}

\title{\Large VideoWeaver: Multimodal Multi-View Video-to-Video \\ Transfer for Embodied Agents
\vspace{-1em}
}

\author{\normalsize George Eskandar$^1$, Fengyi Shen$^1$, Mohammad Altillawi$^1$, Dong Chen$^1$, Yang Bai$^{1, 2, 4}$, Liudi Yang$^{1, 3}$, Ziyuan Liu$^{1, \dagger}$ \\
\normalsize $^1$ Huawei Heisenberg Research Center, $^2$ Ludwig Maximilian University of Munich \\ \normalsize $^3$ University of Freiburg,  $^4$ Munich Center for Machine Learning (MCML)
 }

\begin{document}

\twocolumn[{
  \renewcommand\twocolumn[1][]{#1}
  \maketitle
  \begin{center}
  \vspace{-2em}
    \includegraphics[width=\textwidth]{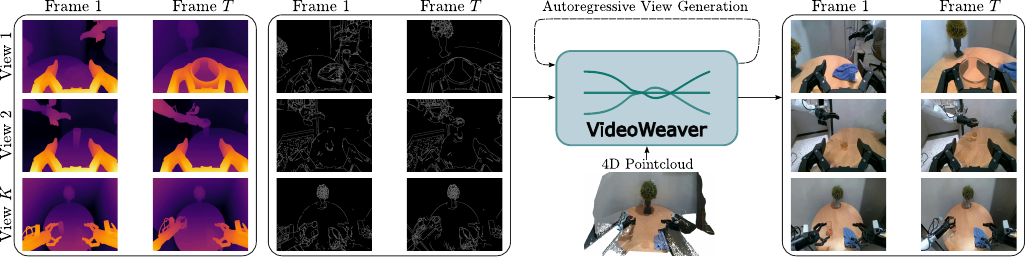}
    \label{fig:teaser}
    \captionof{figure}{We introduce VideoWeaver, the first flow model to synchronously translate multiple viewpoint cameras to a new style. VideoWeaver is a multimodal (depth + sketch) multi-view V2V model that can scale to a large number of views. Our insight is to unify the latent space across views by injecting the coordinates \((x, y, z)\) of a 4D pointcloud (uncolored) inside the flow model.}
  \end{center}
}]

\maketitle
\renewcommand\thefootnote{}\footnotetext{$^{\dagger}$ Corresponding author,  ziyuan.liu@huawei.com
 }
\begin{abstract}
Recent progress in video-to-video (V2V) translation has enabled realistic resimulation of embodied AI demonstrations, a capability that allows pretrained robot policies to be transferable to new environments without additional data collection. However, prior works can only operate on a single view at a time, while embodied AI tasks are commonly captured from multiple synchronized cameras to support policy learning. Naively applying single-view models independently to each camera leads to inconsistent appearance across views, and standard transformer architectures do not scale to multi-view settings due to the quadratic cost of cross-view attention. 

We present VideoWeaver, the first multimodal multi-view V2V translation framework. VideoWeaver is initially trained as a single-view flow-based V2V model. To achieve an extension to the multi-view regime, we propose to ground all views in a shared 4D latent space derived from a feed-forward spatial foundation model, namely, Pi3. This encourages view-consistent appearance even under wide baselines and dynamic camera motion. To scale beyond a fixed number of cameras, we train views at distinct diffusion timesteps, enabling the model to learn both joint and conditional view distributions. This in turn allows autoregressive synthesis of new viewpoints conditioned on existing ones. 

Experiments show superior or similar performance to the state-of-the-art on the single-view translation benchmarks and, for the first time, physically and stylistically consistent multi-view translations, including challenging egocentric and heterogeneous-camera setups central to world randomization for robot learning. 
\end{abstract}
\vspace{-2em}

\section{Introduction}
\label{sec:intro}

Modeling how the physical world evolves is fundamental for training embodied agents. Generalizable policy models~\cite{unipi, unleashinglargescale, pi_0, rt2} often learn to predict plausible actions from video observations, but need vast amount of real-world demonstration data which is expensive to collect. Video generation models~\cite{videodiffmodels} offer to alleviate this need by generating diverse high-fidelity visual data at scale. However, they frequently struggle to maintain physical consistency. As a consequence, a more grounded alternative lies in using Video-to-Video (V2V) translation to learn a mapping from an input video to an output video. Originally introduced in \cite{vid2vid}, V2V conditions the video generation on structured visual controls such as depthmaps, sketches, and semantic segmentation masks. This makes V2V particularly effective for domain randomization: it can re-synthesize simulator or past real-world demonstrations into new styles while preserving the underlying robot action trajectory~\cite{visualparkour, sim_perspectives, simRL, physics_sim, sim_robot}.

In this paper, we move beyond single-view V2V by introducing a formulation that operates across multiple camera viewpoints. We focus on the question: \textit{How can we translate a set of aligned videos of the same scene into a target visual domain while preserving cross-view consistency?} This setting is motivated by recent progress in embodied AI, where multi-camera setups for manipulators~\cite{droid} and humanoid robots~\cite{figure02, unitree_h1, agibot, optimus_gen2} have been shown to improve policy learning~\cite{seefromtheirhands, mvmae, lookcloser, multicamtosinglecam, 3dmvp} due to their richer 3D spatial understanding. However, current V2V models are designed for single-view translation; applying them independently to each camera produces inconsistent view-dependent styles and breaks 3D coherence, limiting their usefulness for multi-view data augmentation.

A number of recent works consider related but distinct problems, including novel view image synthesis~\cite{zero123, zero123pp, mvdream, sv3d, dust3r}, camera-controlled video generation~\cite{seva, cat3d, cameractrl}, novel view synthesis of a dynamic scene~\cite{cameractrl2, cat4d}, and multi-view image edits~\cite{consolidating, coupled_mv_editing, scenecrafter, mvedit, cmd, controllable_ped}. Differently, multi-view V2V does not aim to synthesize unseen viewpoints from available observations, but to jointly translate a set of already captured videos into a target style. However, the underlying 4D representation (space, view, and time) of this task introduces additional constraints on cross-view and temporal consistency.


The main challenge arises from the fact that current video diffusion~\cite{videodiffmodels} and flow~\cite{recflow} models lack mechanisms to enforce 4D consistency, particularly when the cameras are dynamic. Modern robotic platform (i.e. humanoids) employ heterogeneous camera configurations: dynamic first-person cameras (also called ego-centric or in-hand), static first-person cameras (head-mounted) and additional static third-person cameras. These cameras are often widely separated, resulting in limited or even negligible overlap between viewpoints, which weakens classical epipolar and correspondence assumptions. In addition, modeling interactions across all views with standard attention leads to quadratic growth in computation and memory, making naive scaling to more than 3-4 views impractical. Our experiments show that straightforward adaptations from multi-view methods such as adding view-attention layers~\cite{cat3d, mvedit} and camera ray embeddings~\cite{cameractrl, cameractrl2, seva} prove insufficient to maintain a consistent style across all views (see \cref{tab:ablations}). 

\textbf{Contribution.} We present VideoWeaver, the first multi-view V2V flow-based model~\cite{recflow} that can synchronously translate multiple views to a random style defined by a prompt (see Fig. 1). Our main insight is to preserve the underlying 3D world from which spatial-temporal consistency arises, rather than preserving the consistency in the 2D image space. To achieve this, we map all frames across input cameras to a common 4D coordinate system, leveraging recent spatial foundation model like Pi3~\cite{pi3}. The global 4D coordinates are injected into the latent space of our flow model, ensuring a unified representation across views. Architecturally, VideoWeaver is a diffusion transformer~\cite{dit} (DiT) with factorized 4D attention blocks complemented with a Mixture-of-Experts (MoE) module  to condition on depth and sketch controls. 

Moreover, we train the model with different noise timesteps across views, enabling it to learn both the joint distribution over all views and the conditional distribution needed to generate new ones. As a result, VideoWeaver can first synthesize a subset of views and then autoregressively generate additional viewpoints conditioned on those already produced. We summarize our contributions as follows:
\begin{enumerate}[topsep=0pt,itemsep=-1ex,partopsep=1ex,parsep=1ex]
\item We present the first muli-view video translation model for embodied demonstrations conditioned on multimodal spatial controls.
\item We introduce a unified latent space across all views, grounded in a 4D pointcloud extracted from Pi3. This establishes a cross-view spatiotemporal correspondence.
\item We propose to train each view with independent timesteps to model the joint and conditional frames distributions, allowing to autoregressively translate views.
\item VideoWeaver outperforms or is on par to previous V2V models on several evaluation sets from Agibot~\cite{agibot}, Droid~\cite{droid} and Bridge~\cite{bridge}, while maintaining cross-view consistency.  
\end{enumerate}

\section{Related Works}
\label{sec:related_works}

\noindent\textbf{Video-to-Video Translation (V2V)} First V2V approaches were based on Generative Adversarial Networks (GANs) like \cite{vid2vid, worldvid2vid, shortcutv2v, fastv2v, pix2pixHD} but later diffusion-based approaches became more popular due to their training stability and generalization ability. Building on the ControlNet idea for images~\cite{controlnet}, Control-A-Video~\cite{control-a-video} and ControlVideo \cite{controlvideo} extend structural conditioning (edges, depth, pose) to text-driven video editing. VideoComposer~\cite{videocomposer} introduces a set of modality-specific condition encoders and adds the resulting features prior to concatenation with the noisy latents. More recently, VACE  \cite{vace} introduced a multi-task framework via a unified video-condition interface. Cosmos-Transfer-1~\cite{cosmos-transfer-1} trains modality-specific controlnets and only fuses them during inference. However, all these approaches operate on single-view inputs, which is the focus of our multi-view V2V formulation. Moreover, in the single-view case, we introduce a mixture-of-experts module that operates on each spatiotemporal patch to condition on both depth and sketch during training and inference. 

\noindent\textbf{Multi-View Generation.} Most works in this category fall under novel-view synthesis for images~\cite{zero123, zero123pp, mvdream, sv3d, dust3r, cat3d} or for videos~\cite{seva, cat3d, cameractrl, cameractrl2, cat4d, multidiff} which is a different task. However, we use common ideas across these techniques to establish a baseline for our task, notably, camera ray embeddings modules and cross-view attention. More recent 3D-enhanced methods condition a video diffusion model on geometry lifted from a single view, like unprojected depth, to stabilize novel views~\cite{multidiff, uni3c, voyager}. This paradigm is not applicable to our setting: we do not have a single “source image’’ that propagates style to all views as style is defined only by the text prompt in our case, and every view already comes with its own spatial control. Instead, we reconstruct a 4D point cloud using Pi3~\cite{pi3} and inject the resulting coordinates directly into the latent space, without performing 3D-to-2D reprojection.

\noindent \textbf{Multi-View Editing} Another line of work is in multi-view image editing~\cite{consolidating, coupled_mv_editing, scenecrafter, mvedit, cmd, controllable_ped} which has started to gain attention, however these methods do not readily extend to videos. To our knowledge, we are the first to attempt the extension of the V2V task to the multi-view task. 

\noindent\textbf{Editing in 4D.} A closely related field is 4D editing based on Gaussian Splats~\cite{3dgs}. However, State-of-the-art 4D Gaussian Splats (4DGS) methods are impractical for this task, as 4D reconstruction~\cite{4dgs} from a sparse number of views (2-5 views with large distance between the cameras) including moving cameras with motion blur is an ill-posed problem. State-of-the-art 4D editing methods assume a well-reconstructed scene ($>$15 cameras), and take 40 minutes~\cite{efficient4Dedit} to several hours~\cite{instruct4d24d} to edit one scene with 50 frames. Moreover, many methods~\cite{splat4d, 4legs} are limited to single objects. Differently, our method is a flow V2V model~\cite{recflow} that generates multi-view videos of robotic demonstrations (81 frames) conditioned on depth and/or sketch sequences in a few minutes and does not operate in the space of Gaussian Splats.


\section{Method}
\label{sec:method}

\begin{figure*}
    \centering
    \includegraphics[width=\linewidth]{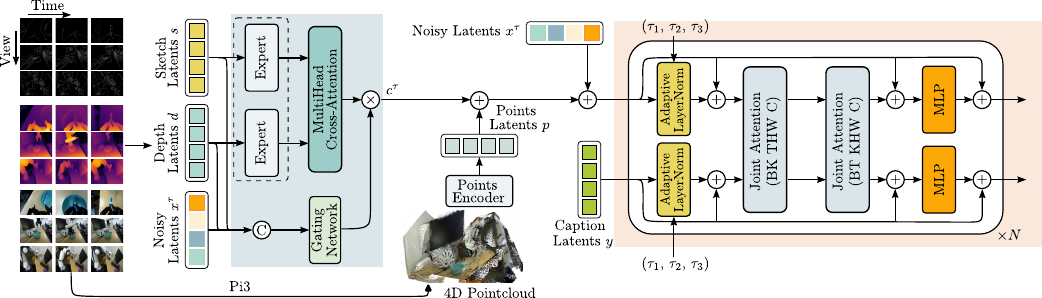}
    \caption{Architecture. VideoWeaver is a DiT with factorized 4D joint attention blocks, and a Mixture-of-Experts module for multimodal conditioning. Moreover, a points encoder conditions the network on a 4D pointcloud estimated from Pi3~\cite{pi3}.}
    \label{fig:method}
    \vspace{-1em}
\end{figure*}
Our aim is to design a flow model that can translate multi-view videos to a style defined by a prompt. We start by formally defining the problem in \cref{sec:problem} and the flow model formulation in \cref{sec:flow_basics}. Our approach can be described in three stages. VideoWeaver is first trained as single-view V2V model conditioned on depth and sketch sequences (\cref{sec:moe}). Then, it is extended as a multi-view V2V model. Key to our method is a unified latent world grounded in a 4D pointcloud, that we present in \cref{sec:sharedlatent}. Then, we show how we can extend the model to a large number of views in \cref{sec:extension}.


\subsection{Problem Definition}
\label{sec:problem}
We consider a scene recorded from $K$ camera viewpoints. For each view $k \in \{1,\dots,K\}$, we are given two spatial controls: a sketch video $s_k \in \mathbb{R}^{1 \times T \times H \times W}$ and a depth video $d_k \in \mathbb{R}^{1 \times T \times H \times W}$. We denote the complete multi-view structural input as $c = \{(s_k, d_k)\}_{k=1}^{K}$ and the text prompt as $y$. The objective is to generate a multi-view RGB video $x = \{x_k\}_{k=1}^{K}$, where each $x_k \in \mathbb{R}^{3 \times T \times H \times W }$ is  aligned to its own spatial controls in $c$ and is geometrically consistent with the other views. We therefore seek to model the  distribution $p_\theta(\{x_k\}_{k=1}^{K} \mid c, y)$, where $\theta$ parametrizes the generative model.

\subsection{Flow-Based Generative Model}
\label{sec:flow_basics}
We adopt a flow-based formulation~\cite{recflow, scalingflow} in which samples are obtained by transporting noise towards the data through a continuous transformation. Let $x^\tau$ denote the state of the sample at timestep $\tau \in [0,1]$, where $x^0 \sim \mathcal{N}(0, I)$ is a standard Gaussian noise sample and $x^1 = x$ corresponds to the final generated video. The evolution of $x^\tau$ is defined by the ordinary differential equation
\begin{equation}
\frac{d x^\tau}{d\tau} = v_\theta(x^\tau, y, \tau),
\end{equation}
where $v_\theta$ is a learnable velocity field conditioned on the multi-view structural input $y$, and is modeled by a DiT. Sampling consists of integrating this Ordinary Differential Equation (ODE) from $\tau=0$ to $\tau=1$ to obtain $x^1$. During training, we follow the rectified flow objective and consider the linear interpolation $x^\tau = (1 - \tau)\,x^0 + \tau\,x^1$ between noise and data. The model is trained to align the velocity field with the displacement direction from noise to data, minimizing
\begin{equation}
\label{eq:flow_loss}
\mathcal{L}(\theta) = \mathbb{E}_{x, y, \tau} \Big[ \big\| v_\theta(x^\tau, y, \tau) - (x - x^0) \big\|^2 \Big].
\end{equation}

\subsection{Single-View V2V Model}
\label{sec:moe}

\noindent\textbf{Architecture.} We construct our V2V model on top of a text-to-video DiT. The model operates in the latent space of a 3D Variational Autoencoder (VAE)~\cite{vae,latentdiff}, which compresses videos by a factor of $8$ along the spatial and temporal dimensions. Each input sketch and depth sequence is therefore encoded into latent feature maps
\[
f_{s},\, f_{d} \in \mathbb{R}^{T \times H \times W \times C}.
\]

\noindent\textbf{Mixture-of-Experts}. To combine sketch and depth information, we introduce a patch-wise Mixture-of-Experts (MoE) fusion module. This design is motivated by the asymmetry of the cues on the patch level: depth often provides reliable geometric structure, but may be uninformative for thin or small objects, while sketch guidance provides stable shape cues but may be ambiguous when edges overlap or foreground and background share similar appearance. Prior approaches typically fuse modalities by addition or concatenation~\cite{videocomposer,cosmos-transfer-1}, forcing the model to treat both signals uniformly. In contrast, our MoE formulation allows the model to adaptively select which modality to rely on at each spatiotemporal location and at each timestep $\tau$.

The MoE module applies two experts $E_s(\cdot)$ and $E_d(\cdot)$, implemented as lightweight convolutional networks, to the two latent streams. Their outputs exchange complementary information using a frame-wise multi-head cross-attention layer with residual connections. A gating network predicts patch-wise mixture weights $\alpha^\tau$ conditioned on the current latent state $x^\tau$, the depth and sketch features $f_{d}$, $f_{s}$ and the timestep $\tau$. The fused structural control signal is given by
\begin{equation}
\label{eq:moe_fusion}
c^\tau = \alpha^\tau \cdot E_s(f_{s}) \;+\; (1 - \alpha^\tau) \cdot E_d(f_{d}).
\end{equation}
This fused control $c^\tau$ is then added to the noisy latents $x^\tau$ at the input of the DiT (see \cref{fig:method}).

\noindent\textbf{Training Details.} Even large foundation models (e.g., VACE~\cite{vace}, Cosmos-Transfer-1~\cite{cosmos-transfer-1}) struggle with first-person views, often producing object flickering or disappearance due to the limited spatial context available in ego-centric footage. To mitigate this, we increase camera-pose diversity in the training data to expose the model to a wider range of ego-centric configurations. Second, we reinforce the learning of global spatial structure. Notably, flow models typically oversample mid-range timesteps~\cite{scalingflow}, which under-exposes the early timesteps that are crucial for establishing the scene layout. We instead adopt uniform timestep sampling, which leads to more stable and consistent multi-view generation. This is likely because early timestep denoising is easier in V2V settings where spatial controls are available. A similar finding was established in \cite{zero123pp}.

\noindent\textbf{Loss function.} To strengthen spatial coherence, particularly at early timesteps, we introduce a wavelet consistency loss on the predicted latents. Motivated by prior work enforcing high-frequency alignment in generative models~\cite{swagan,usis}, we compute a 3D wavelet transform~\cite{wavelet} of the predicted latents $\hat{x}^1$ and the ground-truth video $x$, and minimize their coefficient distance. The predicted latents are obtained by adding back the noise to the model output, i.e., $\hat{x}^1 = x^0 + v_{\theta}(x^{\tau}, y, c^{\tau}, \tau)$. We add this loss to the loss function of \cref{eq:flow_loss}

\subsection{4D Shared Latent Space}
\label{sec:sharedlatent}
To extend our video diffusion model to the multi-view regime, we first equip every frame in every view with camera-ray embeddings, following CameraCtrl~\cite{cameractrl,cameractrl2}, so as to handle moving and view-dependent cameras. On top of this, we insert a cross-view attention layer after each joint (intra-view) attention block. This yields a factorized 4D attention scheme: each DiT block (i) performs standard attention within a view over its $(T \times H \times W)$ tokens, and (ii) subsequently attends at each frame over the $(V \times H \times W)$ spatial tokens across views. In our design, the number of views $K$ is set to 3. 

This design preserves the original single-view weights and enables a smooth architectural extension. Nevertheless, we observe that such factorization alone does not guarantee view-consistent appearance. We find that latent features across views still remain only weakly coupled. Cross-view attention can be easily confused by small or partially visible objects when viewpoints are far apart, and camera-ray embeddings give only coarse geometric cues. 

To enforce a stronger geometric linkage, our strategy is to condition the model on a strong multi-view 4D prior. Recently, VGGT~\cite{vggt} has shown that it is possible to jointly predict camera parameters, depth maps, and a dense pointcloud from a set of input images through a feed-forward network. However, VGGT anchors all predictions to the first view and does not natively support dynamic 4D sequences. 

We therefore adopt Pi3~\cite{pi3}, which is explicitly permutation-invariant over views and reconstructs all observations in a single, affine-invariant global frame. Concretely, we pass all frames ${x_{k,t}}$ across views $k$ and timesteps $t$ to Pi3:
\begin{equation}
\mathcal{F}_{\text{Pi3}}({x_{k,t}}) \rightarrow \bigl( {\hat{K}_k}, {\hat{T}_{k,t}}, \hat{p}_{k,t} \bigr).
\end{equation}
where $\hat{K}_i$ are camera intrinsics, $\hat{T}_i \in SE(3)$ are camera poses, and $\hat{\mathcal{P}}{i}$ are pointclouds.

To the best of our knowledge, prior multi-view video generation works have not exploited Pi3 as a generic 4D conditioning backbone. We therefore investigate whether its permutation-invariant reconstruction can be repurposed for dynamic multi-view sequences, and we find that it indeed provides a stable global frame for our setting (refer to Supplementary).

\begin{figure}
    \centering
    \includegraphics[width=0.9\linewidth]{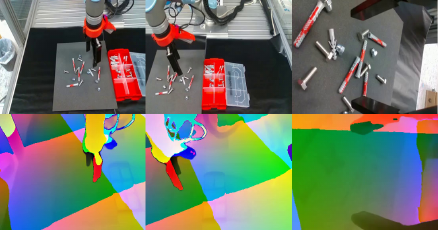}
    \caption{Our insight is to use the output pixelwise pointcloud prediction of Pi3~\cite{pi3} as a correspondence map that unifies the latent space across different views at the input of the DiT. For visualization purposes only, every point \((x, y, z)\) is color-coded to showcase the established spatial correspondences.}
    \label{fig:4Dpointscolored}
    \vspace{-1em}
\end{figure}

Pi3 directly regresses a 3D point per pixel, so the predicted $\hat{\mathcal{P}}_{k,t}$ are already dense and pixel-aligned with the video frames (see \cref{fig:4Dpointscolored}). To make it usable inside the latent space of the flow model, we partition each frame into $8 \times 8$ image patches (matching the VAE downsampling factor) and, for each patch, keep only the point that is closest to the camera, yielding a depth-aware pooling that preserves foreground and contact regions. We further temporally subsample the sequence by keeping every $8$-th frame, which matches the latent stride along time. The resulting low-resolution, temporally thinned 3D grid is passed through a lightweight MLP and injected additively into the corresponding noisy latents.


\subsection{Autoregressive View Generation}
\label{sec:extension}

VideoWeaver natively generates only three views. Extending to \(K>3\) views is not as simple as calling the model repeatedly, because the model as trained so far only estimates the joint distribution \(p_{\theta}(x_1, x_2, x_3 \mid y, c_1, c_2, c_3)\), and does not explicitly learn conditional probabilities such as \(p_{\theta}(x_2, x_3 \mid y, c_2, c_3, x_1)\) or \(p_{\theta}(x_3 \mid y, c_3, x_1, x_2)\), which are exactly the terms we need if we want to autoregressively add views on top of already generated ones.

Our insight is to make the model practice these objectives during training. Inspired by co-generation setups in multimodal diffusion (e.g.\ video and audio co-generation)~\cite{omniflow}, we reinterpret multi-view training as a multi-task problem over paths in the noise-timestep space. In the vanilla setting, the model only sees the path \(\tau: (0, 0 ,0) \rightarrow (1, 1, 1)\), i.e.\ all views share the same timestep and are simultaneously denoised. We generalize this by occasionally freezing one or more views at timestep \(1\) (i.e.\ leaving them clean) and only denoising the remaining ones. Concretely, we introduce paths such as \(\tau: (1, 0 ,0) \rightarrow (1, 1, 1)\) or \(\tau: (0, 1 ,1) \rightarrow (1, 1, 1)\), where views with timestep \(1\) act as already-generated conditioning signals, and the others are denoised in the usual way. In a second training stage, we thus (1) randomly sample a subset of view indices to be treated as ``given'', and set their timestep to \(1\); (2) sample a common timestep for the remaining views and apply noise to them; and (3) mask the loss on the views whose timestep is \(1\), so that no gradients flow from the noised features to the clean features. During inference, when \(K>3\), we first generate the standard three views \((x_1, x_2, x_3)\), then use a subset of them together with the prompt and spatial control of the target view to generate the remaining views in an autoregressive manner.

\section{Experiments}
\label{sec:experiments}

\noindent\textbf{Implementation Details.} We build our V2V model on top of an in-house 11B-parameter text-to-video foundation model, following the Multimodal DiT (MMDiT) architecture~\cite{scalingflow}. Training proceeds in three stages of increasing difficulty: (i) single-view finetuning, to adapt the base model to the V2V setting, (ii) multi-view joint finetuning, where all views are generated synchronously, and (iii) multi-view training with heterogeneous timesteps, enabling the model to learn both joint and conditional view distributions. In each stage, we finetune all parameters using AdamW~\cite{adamw} with a learning rate of $1\mathrm{e}{-4}$. All models can be trained efficiently on 8 Ascend 910B cards (64 GB each) for approximately one week. For inference, we adopt a linear flow scheduler with a discrete Euler solver using 30 integration steps. Generating three synchronized multi-view videos of length 81 frames takes roughly 10 minutes.

\begin{figure*}[!th]
    \centering
    \includegraphics[width=\linewidth]{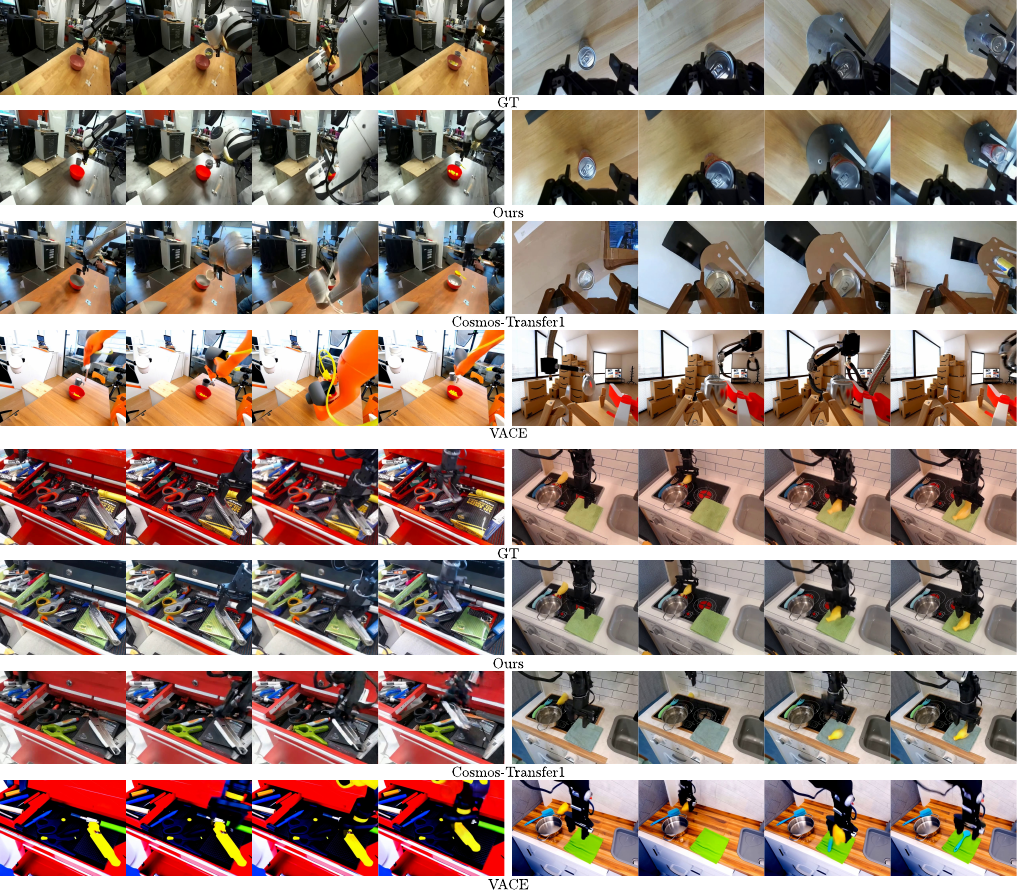}
    \caption{Comparison with state-of-the-art V2V models on single viewpoints from Droid~\cite{droid} and Bridge~\cite{bridge}}
    \label{fig:images_main}
    \vspace{-1em}
\end{figure*}
\noindent\textbf{Datasets.} We train on four datasets covering diverse robotic manipulation settings: Droid~\cite{droid} (140K videos corresponding to two thirds of the dataset), Agibot~\cite{agibot} (75K samples, equivalent to one-third of the dataset), Bridgev2~\cite{bridge} (22K samples, the whole dataset), and an internal dataset of 5K videos. Bridgev2 is only used in the single-view stage, since its multi-view version has a low spatial resolution. This results in almost 240K videos for single-view training and 75K videos for multi-view training.

Droid contains 3–5 cameras per episode (left, right, in-hand, stereo-left, stereo-right); we use the left, right, and in-hand views. Agibot provides three views (left hand, right hand, head). Our internal dataset also has 3 views (left, right, in-hand). In Droid, the in-hand camera is moving, while in Agibot two cameras are moving. For testing, we build a benchmark of 310 single-view samples drawn from Agibot, Droid and Bridgev2, unseen during the training. For the multi-view benchmark, we remove Bridgev2 from the test set, since Bridgev2 has no multi-view samples at high-resolution. Overall, the multi-view test set has 90 samples.

All videos are preprocessed to 81 frames and resized to $480 \times 640$. We keep a single robot action per video. Because the original captions are short and sometimes noisy, we re-caption all videos with Qwen-2.5-VL~\cite{qwen2.5} to describe both the action and the scene. We extract depth using Video-Depth-Anything~\cite{video_depth_anything} and edges using the Canny detector~\cite{canny}. Pi3~\cite{pi3} is used to extract a point cloud at the latent resolution $11 \times 60 \times 80$, and we reuse the camera poses predicted by Pi3 to stay consistent with the injected point cloud.

\begin{figure*}[!th]
    \centering
    \includegraphics[width=\linewidth]{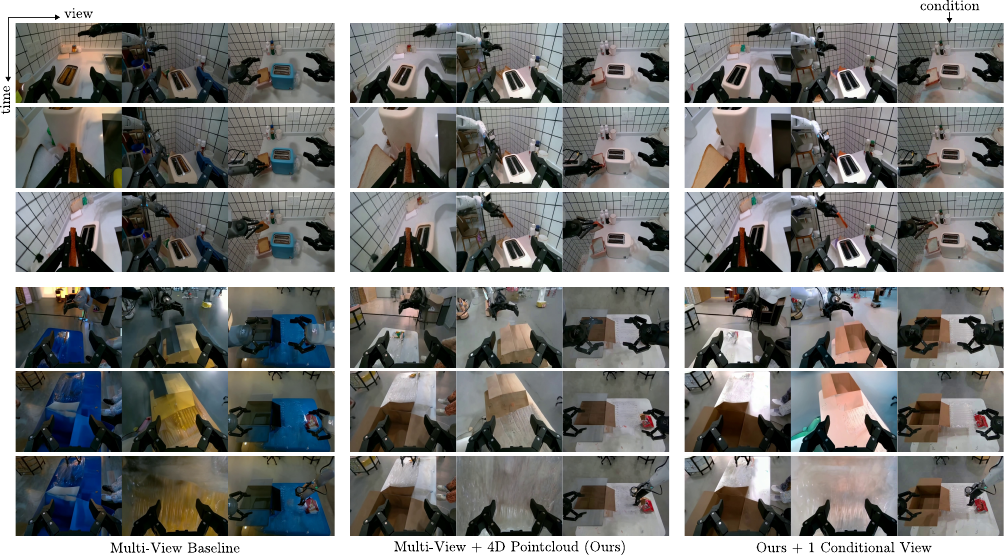}
    \caption{Ablation study of the multi-view model on on Agibot dataset~\cite{agibot}.}
    \label{fig:images_ablations}
\end{figure*}
\begin{table*}
    \centering
    \resizebox{\textwidth}{!}{\begin{tabular}{c | c c c c c | c c c c c| c c c c c}
      \Xhline{1pt}
      \multirow{2}{*}{\textbf{Methods}} & \multicolumn{5}{c |}{Bridge}  &
      \multicolumn{5}{c |}{Droid}  &
      \multicolumn{5}{c}{Agibot} \\
      & VBench$\uparrow$ &  Dover$\uparrow$ & Edge$\uparrow$ & Depth$\downarrow$ & JEDi$\downarrow$ & VBench$\uparrow$ & Dover$\uparrow$ & Edge$\uparrow$ & Depth$\downarrow$ & JEDi$\downarrow$ & VBench$\uparrow$ &  Dover$\uparrow$ & Edge$\uparrow$ & Depth$\downarrow$ & JEDi$\downarrow$ \\ 
      \Xhline{1pt}
      ControlVideo~\cite{controlvideo} & 0.710  & 50.98 & 0.123 & 0.505 & 12.04 & 0.700 & 45.60 & 0.083 & 0.880 & 6.19 & 0.707 & 44.69 & 0.095 & 0.850 & 6.67 \\
      Control-A-Video~\cite{control-a-video} & 0.842 & \textbf{78.24} & 0.137 & 0.276 & 5.16 & 0.687 & 59.13 & 0.086 & 0.610 & 3.20 & 0.70 & 60.31 & 0.09 & 0.619 & 1.65\\
      Cosmos-Transfer1~\cite{cosmos-transfer-1} & \textbf{0.847} & 72.09 & 0.345 & 0.223 & 2.51 & 0.810 & 76.34 & 0.277 & 0.460 & 0.640 & 0.799 & 71.78 & 0.323 & \textbf{0.364} & \textbf{0.22} \\
      VACE~\cite{vace} & 0.833 & 74.93 & 0.135 & 0.258 & 4.39 & 0.729 & \textbf{79.77} & 0.121 & 0.511 & 1.29 & 0.832 & \textbf{72.08} & 0.122 & 0.389 & 0.93 \\
      \Xhline{1pt} 
      Ours Single-View & 0.837 & 74.01 & \textbf{0.393} & \textbf{0.158} & \textbf{1.67} & 0.732 & 59.96 & 0.359 & 0.362 & \textbf{0.384} & 0.713 & 66.69 & 0.373 & 0.468 & 0.23 \\
      Ours Multi-View & N/A & N/A & N/A & N/A & N/A & \textbf{0.736} & 60.02 & \textbf{0.376} & \textbf{0.347} & 0.509 & 0.720 & 67.34 & \textbf{0.378} & 0.394 & \textbf{0.22}  \\
      \Xhline{1pt}
    \end{tabular}}
    \caption{ Comparison against state-of-the-art V2V baselines on the test set. We show detailed results for Droid, Agibot and Bridgev2. Edge and Depth are short for Edge-F1 score and Depth-siRMSE.}
    \label{tab:main_comparisons}
    \vspace{-1em}
\end{table*}

\noindent\textbf{Baselines.} Since there is no baseline for multi-view V2V model, we compare with the best available single-view V2V methods with open-source code. Namely, we choose 4 baselines: 
\begin{itemize}
    \item VACE~\cite{vace} is a 14B parameters model trained on a large scale dataset containing various tasks, including depth-to-video. VACE is a ControlNet-based model.
    \item Cosmos-Transfer-1~\cite{cosmos-transfer-1}, a 7B parameter model, trained on a large scale dataset of Embodied AI and autonomous driving videos. Cosmos-Transfer-1 trains a Controlnet separately for each modality and fuses them at test-time. During evaluation, we use their depth and sketch ControlNets, giving them equally a weight of $0.5$.
    \item ControlVideo~\cite{controlvideo} is another depth-to-video ControlNet method based on Stable Diffusion~\cite{latentdiff}.
    \item Control-A-Video~\cite{control-a-video} uses Controlnet and motion priors on top of a latent diffusion model~\cite{latentdiff} to achieve depth-to-video translation.
\end{itemize}
We run the baselines on each view independently, using their official inference scripts. As a multi-view baseline, we use our foundation model to which we insert cross-view attention layers and camera ray embeddings~\cite{cameractrl, cameractrl2} at the start. Moreover, we compare with our single-view model.

\noindent\textbf{Evaluation Metrics.} In the single-view setting, we assess three aspects: (i) alignment with the spatial control, (ii) video quality and temporal consistency, and (iii) realism. For alignment, following Cosmos-Transfer-1~\cite{cosmos-transfer-1}, we re-extract depth and sketch from the generated video and report (a) siRMSE between the predicted and ground-truth depth, and (b) Edge-F1 between the predicted and ground-truth sketch. For video quality, we use VBench~\cite{vbench} and compute its six dimensions (Subject Consistency, Background Consistency, Motion Smoothness, Dynamic Degree, Aesthetic Quality, Imaging Quality) and average them into a single score. We also report the Dover score~\cite{dover}, a neural-network-based measure of aesthetic and technical quality. For realism, we use the JEDi metric~\cite{jedi}, which has been shown to be more reliable than FVD~\cite{fvd}. For the multi-view setting, we additionally report Met3R~\cite{met3r}, which measures cross-view consistency using DUST3R~\cite{dust3r} and DINOv2~\cite{dinov2} features. We apply Met3R on every pair of views across all frames and average the scores.

\subsection{Main Comparisons}
\label{main_comparisons}
In \cref{tab:main_comparisons}, we compare our single-view and multi-view models against several V2V baselines. VideoWeaver outperforms prior work in most settings, including VACE—which uses more parameters and training data—and Cosmos-Transfer-1, trained on a larger Physical-AI dataset. On Droid and Agibot, the multi-view variant even improves over the single-view model thanks to the additional learned consistency. Overall, our strengths lie in alignment and visual fidelity, whereas we trail slightly on the Dover metric and the aesthetic score of VBench. This may stem from the strong temporal downsampling (factor 8) in our VAE, which can introduce mild blur but enables training on longer clips.

In \cref{fig:images_main}, we present qualitative results on Bridge and Droid. Our model produces more consistent interactions from diverse viewpoints, including first-person views. VACE achieves high aesthetic quality but occasionally exhibits unnatural textures, likely due to synthetic data in the training, whereas our model is only trained on real data.

\subsection{Multi-View Ablations}
\label{mv_ablations}
In \cref{tab:ablations}, we first compare our method to a straightforward baseline built directly on top of the foundation model. Following prior work in novel-view synthesis~\cite{cameractrl, cameractrl2} and 3D scene generation~\cite{cat3d, cat4d}, this baseline augments the model with camera-ray embeddings and multi-view attention blocks. As shown in \cref{fig:images_ablations}, this design struggles to maintain cross-view coherence: the same object frequently appears with inconsistent colors.

By contrast, unifying the latent space through our 4D point-cloud injection yields immediate and substantial improvements. Colors remain consistent across the generated views, and the Met3R score on the Droid dataset increases by $10\%$. While this gain is less pronounced numerically on the Agibot dataset, the qualitative improvements are clear (more visualizations in the Appendix).

To evaluate whether our heterogeneous-timestep training strategy produces measurable benefits, we also conduct a conditional-consistency test: we randomly select one view as a conditioning input, generate the remaining two views, and average the results over all three conditioning permutations. As reported in \cref{tab:ablations}, this strategy yields a clear improvement in Met3R, indicating stronger cross-view agreement. In \cref{fig:images_ablations}, the resulting images exhibit  sharper geometry and stable appearance across viewpoints.

\begin{table}
    \centering
    \resizebox{\linewidth}{!}{\begin{tabular}{l | c | c }
      \Xhline{1pt}
      \multirow{2}{*}{\textbf{Methods}} & \multicolumn{1}{c |}{Agibot Test Set} & \multicolumn{1}{c}{Droid Test Set}  \\
      & Met3R$\uparrow$ & Met3R$\uparrow$   \\ 
      \Xhline{1pt}
      Multi-View Baseline (Camera Ray Embedding) & 0.597 & 0.481  \\
      Multi-View + 4D Pcd & 0.612 & 0.533 \\
      Cond. Multi-View + 4D Pcd (1 View) & \textbf{0.624} & \textbf{0.578}   \\
      \Xhline{1pt}
    \end{tabular}}
    \caption{Ablations Study of our model on the multi-view test set. Pcd denotes the pointcloud, Cond. Multi-View is the conditional model trained with heterogeneous timesteps. }
    \label{tab:ablations}
    \vspace{-1em}
\end{table}
\subsection{Single-View Ablations}
\label{sv_ablations}
Finally, in \cref{tab:sv_ablations}, we evaluate the impact of our MoE module by training two ablated variants—one using depth only and one using sketch only—while keeping the training setup of \cref{sec:moe}. Overall, sketch performs better on robotic demonstration data, as depth can be ambiguous in cluttered lab environments. The MoE effectively leverages both modalities and yields more consistent videos that remain aligned with both conditions.

Additionally in the last two rows, we show that one modality can be dropped during inference with minimal effect on the evaluation metrics. This is due to both the MoE design and the training strategy where we occasionally drop one modality with a low probability.

\subsection{Generalization on Unseen Datasets}
\label{generalization}
VideoWeaver exhibits remarkable generalization across unseen tasks, viewpoints, embodiments, and environments. In the supplementary material, we show the performance on three datasets, namely Berkeley-Autolab \cite{BerkeleyUR5Website}, Robomind \cite{Robomind} and a private simulator dataset. All three were excluded from training. Additionally, we provide results with diverse style randomization across multiple views, achieved by conditioning on different prompts.

\begin{table}
    \centering
    \resizebox{\linewidth}{!}{\begin{tabular}{l | c c c c c}
      \Xhline{1pt}
      \multirow{2}{*}{\textbf{Methods}} & \multicolumn{5}{c}{Bridge Test Set}  \\
      & VBench$\uparrow$ &  Dover$\uparrow$ & Edge$\uparrow$ & Depth$\downarrow$ & JEDi$\downarrow$ \\ 
      \Xhline{1pt}
      Sketch-to-Video & 0.833 & 72.01 & \textbf{0.394} & 0.245 & \textbf{1.67}  \\
      Depth-to-Video & 0.764 & \textbf{74.51} & 0.250 & 0.199 & 2.03   \\
      Ours with MoE &\textbf{ 0.837} & 74.01 & 0.393 & \textbf{0.158} & \textbf{1.67}   \\
      \midrule[1.5pt]
      Ours with MoE- Drop Depth & 0.829 & 70.41 & 0.393 & 0.159 & 1.58   \\
      Ours with MoE- Drop Sketch & 0.833 & 69.70 & 0.171 & 0.226 & 1.79    \\
      \hline
      \Xhline{1pt}
    \end{tabular}}
    \caption{Ablations Study of our model on the single-view test set.  }
    \label{tab:sv_ablations}
    \vspace{-1em}
\end{table}

\subsection{Limitations}
Our approach still has a few practical limitations. First, some inconsistencies persist, especially for small objects, likely due to the required downsampling of the pointmap to match the latent resolution. Second, although we can autoregressively generate additional views, the model is limited to a fixed number of frames and cannot natively produce arbitrarily long rollouts. In particular, it lacks a temporal autoregressive mechanism to extend the multi-view sequence. A natural direction is to incorporate temporal self-conditioning methods that reuse previously generated frames as context, such as~\cite{diffusionforcing, selfforcing}.

\section{Conclusion}
\label{sec:conclusion}

In this paper, we introduce VideoWeaver, the first framework for multimodal, multi-view video translation tailored to Embodied AI. At its core, VideoWeaver leverages a 4D point-cloud representation predicted in a feed-forward manner by a spatial foundation model, enabling us to unify the latent space across views and preserve a consistent underlying 3D structure. In addition, our heterogeneous timestep training strategy allows to learn both joint and conditional distributions, providing a principled mechanism for extending the generated views beyond the input set.

Through extensive experiments on three challenging Embodied AI benchmarks—Droid, Agibot, and Bridge—we demonstrate strong generalization capabilities, high cross-view consistency, and robust handling of diverse camera configurations. Beyond video translation, our formulation opens the door to extending this framework to broader vision tasks that benefit from multi-view reasoning, such as 4D scene generation, multi-camera perception for robotics, and view-consistent video editing.

\newpage
{
    \small
    \bibliographystyle{ieeenat_fullname}
    \bibliography{main}
}

\clearpage
\setcounter{page}{1}
\maketitlesupplementary

This supplementary is structured in the following way. In \cref{sec:suppl_pcd}, we show visualizations of pointclouds from Pi3. Finally, in \cref{sec:qual}, we show a sample of generated videos from our model. For a more extensive visualizations, please refer to the video supplementary. 

\section{4D Pointcloud from Pi3}
\label{sec:suppl_pcd}

\begin{figure*}
    \centering
    \includegraphics[width=0.9\linewidth]{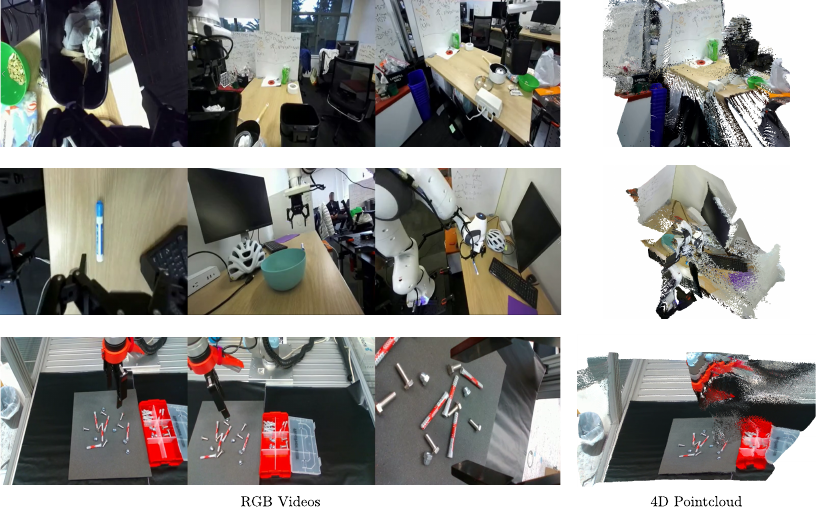}
    \caption{We show several visualizations of the estimated 4D pointcloud from Pi3 and some of their corresponding RGB frames.}
    \label{fig:suppl_4D}
    \vspace{-1em}
\end{figure*}

We visualize several estimated point clouds obtained from multi-view videos in \cref{fig:suppl_4D}. Leveraging its permutation-invariant design, Pi3 can reconstruct the underlying scene in 4D. To do so, we sample the same number of frames from each viewpoint and provide all frames to Pi3 simultaneously, without requiring any specific ordering.  

\section{Generalization to Unseen Datasets}
\label{sec:suppl_results}

\begin{figure*}[!ht]
    \centering
    \includegraphics[width=0.95\linewidth]{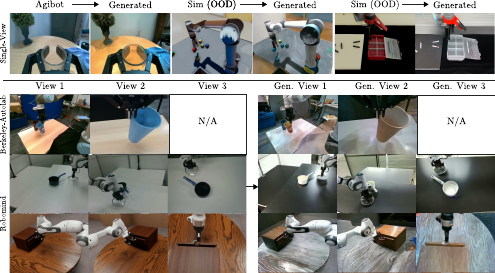}
    \caption{Inference on novel datasets, Berkeley, Robomind and an internal simulator data with reflective and transparent surfaces. }
    \label{fig:rebuttal}
    \vspace{-1em}
\end{figure*}

We show out-of-distribution results on a single-view simulator dataset and two multi-view real datasets (Berkeley-Autolab and Robomind) in Figure \ref{fig:rebuttal}. Berkeley-Autolab has only two views. The three datasets present different tasks, embodiments (robot types), camera poses and texture (synthetic data included) unseen during training. VideoWeaver exhibits strong generalization ability across these scenarios. However, failure cases  still happen and include mostly cluttered scenes or small objects in normal environments. 

\section{Text Prompts}
\label{sec:suppl_caption}
We re-caption all videos using Qwen-2.5-VL~\cite{qwen2.5}. For multi-view videos, we select the viewpoint that provides the clearest and most comprehensive visibility of the environment. The same caption strategy is applied to all videos: we first caption the action performed by the robot "[Action] A white and black robotic arm is holding a small container and pouring liquid into a bright red bowl on the table. [Description] The robotic arm has a white exterior with black joints and is positioned over a dark brown table. The bowl is bright red and sits near the center of the table. Nearby, there are small objects, including a light-colored cylindrical item and a thin tool. To the left, there is a light wooden box next to a black cabinet or machine. In the background, there are desks, chairs, and various equipment in black, gray, and metallic colors, along with visible cables and lab tools." For style optimization, we keep the same caption structure and only change how the objects look like. Since we only train on real data, artistic styles are not supported. When the caption is too long, or the scene is too cluttered, the prompt following is weakened. 

\section{Additional Qualitative Results}
\label{sec:qual}

\begin{figure*}[!th]
    \centering
    \includegraphics[width=\linewidth]{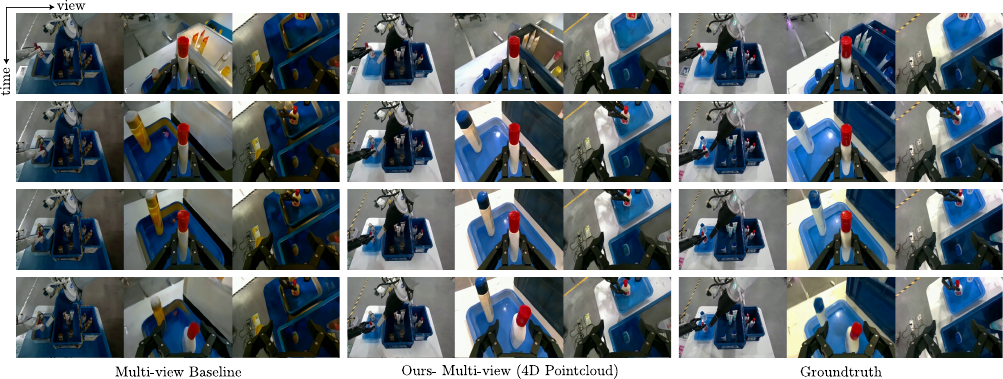}
    \caption{Additional qualitative comparisons between our full multi-view model and its ablated counterparts.}
    \label{fig:qual_suppl}
    \vspace{-1em}
\end{figure*}

\begin{figure*}[!th]
    \centering
    \includegraphics[width=\linewidth]{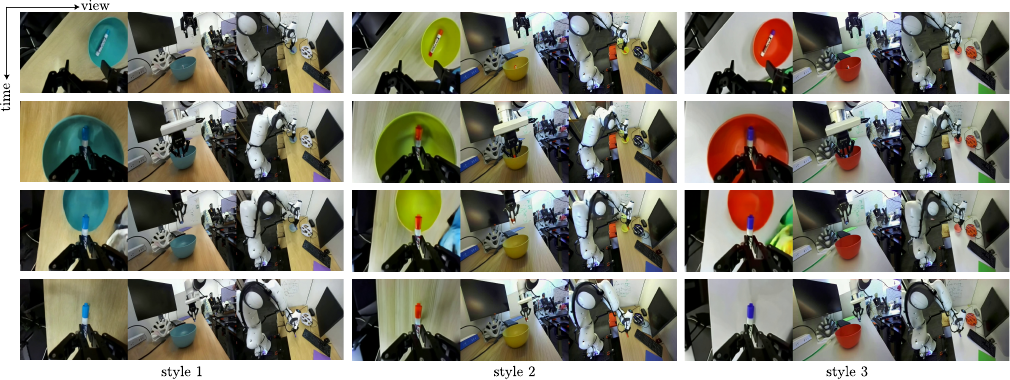}
    \caption{Additional qualitative results generated from our multi-view model by varying the prompt.}
    \label{fig:random_suppl}
\end{figure*}

We show more qualitative results of the multi-view model with and without the 4D pointcloud injection from Pi3 in \cref{fig:qual_suppl}. Moreover, in \cref{fig:random_suppl}, we show the ability of our model to generate diverse styles of the same multi-view video only by varying the text prompt (depth, sketch and pointcloud remain the same). VideoWeaver can perform localized multi-view edits in challenging viewpoints. More extensive viusalizationsare shown in the video supplementary.

\end{document}